\documentclass[letterpaper, 10 pt, conference]{ieeeconf}  %

\IEEEoverridecommandlockouts                              %

\usepackage{times}
\usepackage{graphics} %
\usepackage{graphicx}
\usepackage{subfigure}
\usepackage{amsmath,amssymb,amsopn,amstext,amsfonts}
\usepackage{cancel}
\usepackage[space]{cite}
\usepackage{pdfsync}
\usepackage{balance}
\usepackage{color}
\usepackage{mathtools}
\usepackage{bm}

\usepackage{diagbox}
\usepackage{float}
\usepackage{epstopdf}
\usepackage{pifont}
\usepackage{fixltx2e}
\usepackage{amsmath}
\usepackage{multirow}
\usepackage{url}
\usepackage{verbatim}
\usepackage{caption}
\usepackage{adjustbox}
\usepackage{booktabs}
\usepackage{threeparttable}
\usepackage{makecell}
\usepackage{upgreek}
\usepackage{algorithm, algpseudocode}

\makeatletter
\let\NAT@parse\undefined
\makeatother
\usepackage[linkcolor=red,citecolor=green,urlcolor=red,colorlinks=true]{hyperref}
\title{\LARGE \bf Spatio-Temporal Calibration for Omni-Directional Vehicle-Mounted Event Cameras}
\author{Xiao Li$^{1}$*, Yi Zhou$^{2}$*, Ruibin Guo$^{1}$, Xin Peng$^{3}$, Zongtan Zhou$^{1}$, Huimin Lu$^{1}$%
\thanks{${1}$ College of Intelligence Science and Technology, National University of Defense Technology. {\tt\small lhmnew@nudt.edu.cn} }
\thanks{${2}$ Neuromorphic Automation and Intelligence Lab (NAIL) at the School of Robotics, Hunan University. {\tt\small eeyzhou@hnu.edu.cn}}
\thanks{${3}$ Motovis Intelligent Technologies (Shanghai) Co Ltd.}
\thanks{* denotes equal contribution. 
Corresponding author: Yi Zhou and Huimin Lu.}
\thanks{This work was supported in part by the National Science Foundation of China under Grant U1913202, U22A2059, and 62203460, as well as Major Project of Natural Science Foundation of Hunan Province under Grant 2021JC0004.}
}

\global\long\def\bfr{\mathbf{r}}
\global\long\def\bfs{\mathbf{s}}
\global\long\def\bft{\mathbf{t}}

\global\long\def\bfv{\mathbf{v}}

\global\long\def\bfA{\mathbf{A}}
\global\long\def\bfB{\mathbf{B}}

\global\long\def\bfH{\mathbf{H}}
\global\long\def\bfI{\mathbf{I}}

\global\long\def\bfU{\mathbf{U}}
\global\long\def\bfV{\mathbf{V}}

\global\long\def\bfX{\mathbf{X}}

\global\long\def\ttA{\mathtt{A}}
\global\long\def\ttB{\mathtt{B}}

\global\long\def\ttE{\mathtt{E}}

\global\long\def\ttT{\mathtt{T}}

\global\long\def\cV{\mathcal{V}}

\global\long\def\Rot{\mathtt{R}}

\global\long\def\etal{\textit{et~al.}}
\begin{document}

\maketitle
\thispagestyle{empty}
\pagestyle{empty}

\begin{abstract}
We present a solution to the problem of spatio-temporal calibration for event cameras mounted on an onmi-directional vehicle.
Different from traditional methods that typically determine the camera's pose with respect to the vehicle's body frame using alignment of trajectories, our approach leverages the kinematic correlation of two sets of linear velocity estimates from event data and wheel odometers, respectively.
The overall calibration task consists of estimating the underlying temporal offset between the two heterogeneous sensors, and furthermore, recovering the extrinsic rotation that defines the linear relationship between the two sets of velocity estimates.
The first sub-problem is formulated as an optimization one, which looks for the optimal temporal offset that maximizes a correlation measurement invariant to arbitrary linear transformation.
Once the temporal offset is compensated, the extrinsic rotation can be worked out with an iterative closed-form solver that incrementally registers associated linear velocity estimates.
The proposed algorithm is proved effective on both synthetic data and real data, outperforming traditional methods based on alignment of trajectories.
\end{abstract}
\vspace{0.2cm}

\section*{Multimedia Material}
\noindent Code: {\small\url{https://github.com/esheroe/EvCalib.git}}\\

\section{Introduction}
\label{sec: introduction}

Extrinsic calibration is a prerequisite to almost any mobile robot application, because measurements from different sensor modalities are sometimes processed and even fused into a unified coordinate system, such as a robot's body frame.
For an autonomous ground vehicle equipped with cameras, extrinsic calibration refers to the operation that determines each camera's mounting position and orientation with respect to the vehicle's body frame.
Existing solutions designed for standard cameras typically run a pipeline based on alignment of trajectories, which estimates the extrinsic parameters by registering two trajectories recovered from vision information and wheel odometers, respectively.
This widely used approach is, however, inapplicable when the recovered trajectory from the vision end is unreliable.

Different from its standard counterpart, an event camera is a biologically-inspired novel sensor which reports only brightness changes asynchronously.
This unique characteristic leads to better performance in terms of temporal resolution and dynamic range.
Thus, event cameras are suitable for dealing with robotic perception \cite{Rebecq18ijcv, Zhou18eccv, Zhu18eccv, stoffregen2019event, zhou2021emsgc}, localization \cite{Kim16eccv,Rebecq17ral,Rosinol18ral,zhou2021event} and control \cite{Conradt09iscas,Delbruck13fns} tasks involving aggressive motion and high-dynamic-range (HDR) illumination conditions.
However, existing image processing techniques designed for standard vision cannot be applied straightforwardly to event data due to the special output format.
Specifically, mature feature detection and matching techniques, based on which reliable and long-term event data association is to be established and maintained, are lacking.
Hence, techniques that can effectively eliminate accumulated errors in the recovered trajectory, e.g., local bundle adjustment (BA) \cite{triggs2000bundle}, are still not available in event-based visual odometry.
Consequently, the drifted trajectory will lead to inaccurate extrinsic calibration results when applying trajectory alignment based pipelines.
This issue poses a challenge when trying to extrinsically calibrate an event camera to an acceptable level.

\begin{figure}[t]
\centering
\includegraphics[width=0.75\linewidth]{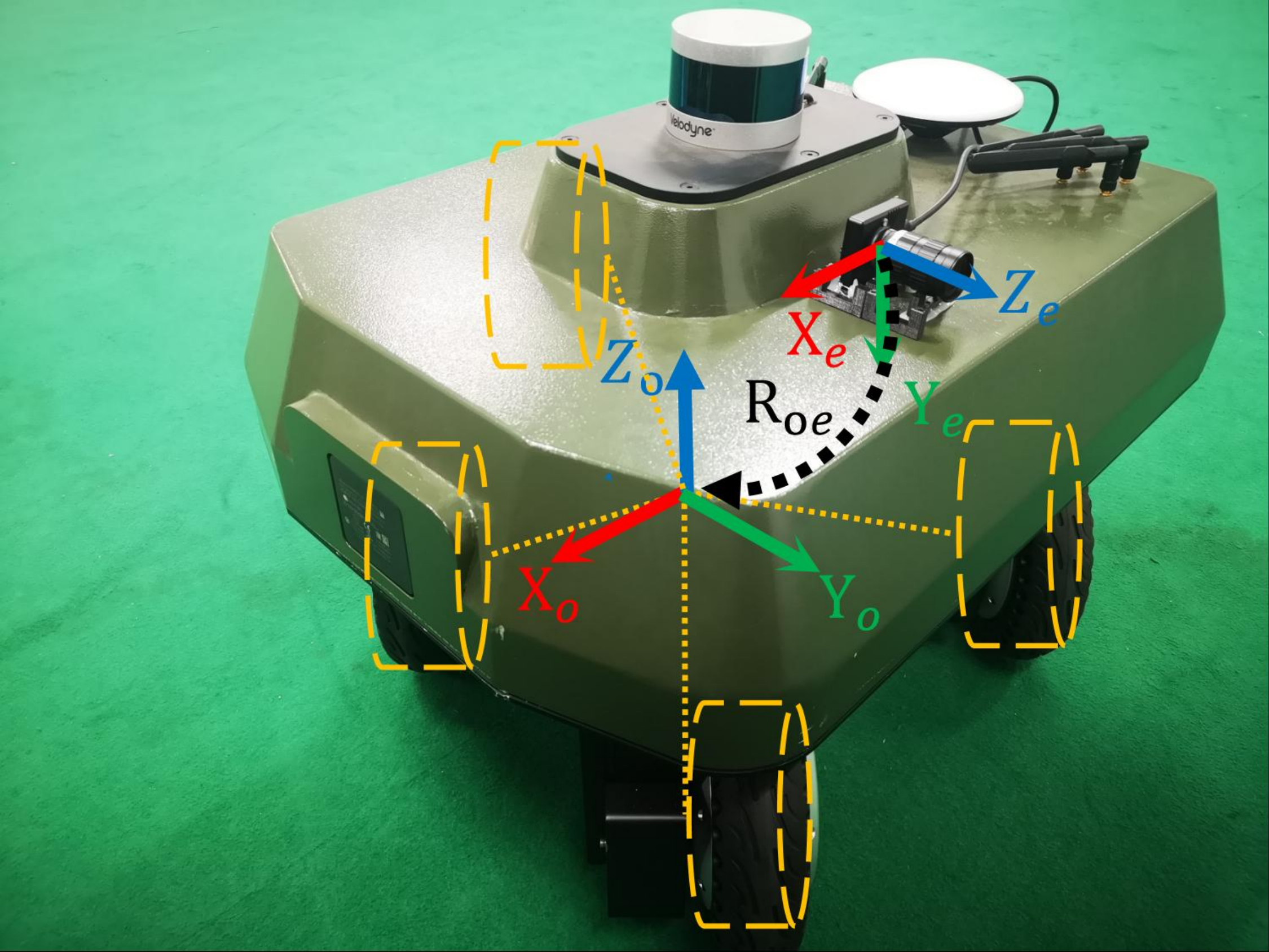}
\caption{Illustration of the geometry of the calibration problem. 
The goal is to recover the relative orientation ($\Rot_{oe}$) of the event camera with respect to the mobile platform body frame, while having the temporal offset between the two heterogeneous sensors (i.e., the odometers and the event camera) compensated.}
\label{fig:eye catcher}
 \vspace{-0.9cm}
\end{figure}

In this paper, we look into the problem of extrinsic calibration of event cameras mounted on an omni-directional mobile platform, as shown in Fig.~\ref{fig:eye catcher}.
Omni-directional vehicles, including non-holonomic all-wheel-steering vehicles and holonomic omni-directional vehicles, are mobility systems that are capable of moving in arbitrary direction.
Such a flexible and potentially aggressive maneuverability leads to a great need for event-based vision.
To circumvent the issue of inaccurate trajectory estimates from an event-based visual odometry, we propose a novel calibration method that recovers the temporal offset and the spatial extrinsics by exploiting the kinematic correlation between linear velocity estimates obtained from event data and wheel odometers, respectively.
The core of our method is a Canonical Correlation Analysis (CCA) process, which evaluates the underlying linear correlation between the two sets of linear velocity estimates.
The applied trace correlation measurement is invariant to arbitrary linear transformation, and thus, the temporal offset can be recovered by maximizing the correlation measurement between the two sets of spatially non-aligned data.
With the temporally-aligned data, the extrinsic rotation can be furthermore worked out with an iterative closed-form solver that incrementally registers associated linear velocity estimates.

The contribution of this paper consists of the following aspects:
\begin{itemize}
    \item An efficient and robust method that determines the direction of linear velocity of a moving event camera, using as input the short-term feature correspondences obtained from the proposed speed-invariant image representation of event data.
    \item A rotation-free solution to the problem of spatio-temporal calibration for omni-directional vehicle mounted event cameras, with special consideration to the mobility system's kinematic characteristics and the limited accuracy of event-based visual odometry. 
    We leverage the kinematic correlation between the linear velocity estimates from two heterogeneous sources, and build the solver on top of a CCA scheme.
    \item An extensive evaluation of the proposed method using both synthetic and real data, and an open-source implementation of our method.
\end{itemize}

The paper is organized as follows.
A literature review on methods of cameras' extrinsic calibration is provided in Section~\ref{sec: related work}.
The proposed method is detailed in Section~\ref{sec:method}, followed with experimental evaluation and relevant analysis in Section~\ref{sec:evaluation}.
We draw the conclusion in Section~\ref{sec:conclusion}.
\section{Related Work}
\label{sec: related work}

\subsection{Trajectory Alignment based Methods} 
\label{subsec: Trajectory Alignment based Methods}
The main category of extrinsic calibration methods is hand-eye calibration~\cite{Tsai89tro, horaud1995hand, strobl2006optimal}, which is based on aligning multiple pose estimates (i.e., trajectories) from two independent sensors.
The hand-eye calibration method establishes the extrinsic constraint using a closed loop of local transformations:
\begin{equation}
    \bfA \bfX = \bfX \bfB,
\label{eq: hand-eye calibration}    
\end{equation}
where $\bfA$ and $\bfB$ represent, respectively, the poses of the two involved sensors at the same time instant, and $\bfX$ the unknown extrinsic parameters, all in homogeneous form $\left( \begin{matrix}    \Rot & \bft \\ \mathbf{0}^{\ttT} & 1 \\ \end{matrix} \right)$.
The hand-eye calibration pipeline typically consists of two steps: 1) Obtaining two sets of time-synchronized pose estimates; 2) Looking for the optimal extrinsic parameters that align maximally the two trajectories.
Given time-synchronized poses $\xi_{\ttA} = \{ \bfA_0, \bfA_1, \dots, \bfA_N \}$ and $\xi_{\ttB} = \{ \bfB_0, \bfB_1, \dots, \bfB_N \}$, the trajectory alignment task is accomplished by solving a non-linear optimization problem with an objective function:
\begin{equation}
    \Rot_{\ttE}^{\ast}, \bft_{\ttE}^{\ast} = \arg\max_{\Rot_{\ttE}, \bft_{\ttE}} \sum_{i=0}^N \Vert \bfA_{i} \bfX - \bfX \bfB_{i} \Vert,
\label{eq: hand-eye calibration energy function}
\end{equation}
where $\Rot_{\ttE}$ and $\bft_{\ttE}$ represent the under estimated extrinsic parameters. 

The hand-eye calibration method and its variants haven been witnessed in many works that estimate the extrinsics between two heterogeneous sensors, especially those recovering the pose of an exteroceptive sensor with respect to the body frame of a robot.
Censi \etal~\cite{Censi2013tro} utilize hand-eye calibration to calculate the extrinsic pose of a range finder with respect to a differential-drive robot's body frame.
The work is carried out without knowing the odometer parameters (i.e., wheel radii and distance between wheels) as a prior, and thus, it jointly estimates the intrinsic odometry parameters and the extrinsic sensor pose.
Assuming all intrinsics are known, Guo \etal~\cite{Guo12icra} unlock the limitation that original hand-eye calibration cannot recover all the three degrees of freedom (DoF) in the camera's orientation.
Their method formulates a least-squares problem to estimate a subset of the odometer-camera rotation parameters, and furthermore, uses these parameters to formulate a second least-squares problem for estimating the remaining unknown parameters of the odometer-camera transformation.
Besides, some globally optimal solutions based on hand-eye calibration, such as \cite{Wodtko21ic3dv, Horn2021ral, giamou2019certifiably}, are developed with special consideration in parametrization of motion parameters.

The accuracy of hand-eye calibration and its variants is largely up to the quality of the input sensor poses in terms of temporal synchronicity and spatial consistency.
To improve temporal synchronicity between input signals, the underlying temporal offset between the two sensors needs to be considered as an additional variable to be optimized in Eq.~\ref{eq: hand-eye calibration energy function}.
Consequently, the temporal offset and the extrinsic parameters are solved jointly, and thus, a proper initial value is needed.
However, hand-eye calibration is not suitable for spatio-temporal calibration of event cameras, because state-of-the-art event-based visual odometry methods \cite{Kim16eccv, Rebecq17ral, zhou2021event} cannot suppress drifts in the recovered trajectory effectively.

\subsection{Motion Correlation based Methods}
\label{subsec: Motion Correlation based Methods}

Different from trajectory alignment based methods, motion correlation based methods can decouple the overall calibration task into sub problems of recovering temporal offset and spatial transformation.
The sub problem of recovering temporal offset can be solved using cross correlation, which is widely used in signal processing \cite{azaria1984time,fertner1986comparison}.
The optimal temporal offset $\delta t^{\ast}$ can be found by maximizing the following objective function:
\begin{equation}
    \delta t^{\ast} = \arg\max_{\delta t} \sum_{i=0}^{N} {}^{a}\zeta(t_i + \delta t) \cdot {}^{b}\zeta_{} (t_i),
\label{eq: cross correlation}    
\end{equation}
where ${}^{a}\zeta$ and ${}^{b}\zeta$ represent the identical 1-D signal measured by sensor $a$ and $b$, respectively.
This cross correlation can be established on any kinematic measurements (or estimates) independent of coordinate system, such as the absolute angular velocity, which has been used to solve the task of IMU-to-Camera temporal calibration~\cite{mair2011spatio}.
The limitation of the cross-correlation pipeline (Eq.~\ref{eq: cross correlation}) is that it is hardly extended to high-dimensional data that are frame dependent.
Consequently, the recovered temporal offset is likely to be inaccurate when data are biased in the missing degrees of freedom.
To overcome this limitation, Qiu \etal~\cite{qiu2020real} propose a unified calibration framework based on 3-D motion correlation, which can efficiently work out the temporal offset and extrinsic rotation between two heterogeneous sensors.
The method can evaluate the correlation of two sets of multi-variate random vectors (e.g., 3-D angular velocity), because the applied CCA technique \cite{thompson2000canonical} is invariant to the underlying (unknown) linear transformation between the two kinematic signals.
Although 3-D angular velocity can be estimated from event data \cite{Gallego17ral}, the 3-D motion correlation on angular velocity estimates is, however, inapplicable to our task, because one DoF of the 3D orientation is never observable.
Fortunately, we could leverage the omni-directional locomotion property of holonomic vehicles that can generate pure translation in arbitrary directions (within the ground plane).
Therefore, our method takes advantage of the correlation between linear velocity estimates from heterogeneous sensors and it is also built on top of the CCA scheme.
To estimate the direction of linear velocity (heading direction) from event data, we determine the epipolar geometry using as input the feature correspondences obtained on a novel speed-invariant image representation of event data.

\section{Methodology}
\label{sec:method}
Given two sets of linear velocity estimates from an event camera and odometers respectively, the goal is to recover the temporal offset between the two heterogeneous sensors, and furthermore, determine the orientation of the event camera with respect to the vehicle's body frame.
In this section, we first discuss our method for estimating the direction of linear velocity from pure event data (Sec.~\ref{subsec:Determining Heading Direction from Event Data}).
Second, we demonstrate the way of acquiring body-frame velocity according to the kinematic model of the omni-directional vehicle used (Sec.~\ref{subsec:Acquiring Body-Frame Velocity from Kinematic Model}).
Finally, we disclose the method that recovers the temporal offset and spatial extrinsics by maximizing the kinematic correlation between linear velocities estimates from heterogeneous sources (Sec.~\ref{subsec:Extrinsic Calibration via Correlation Maximization}).

\subsection{Determining Heading Direction from Event Data}
\label{subsec:Determining Heading Direction from Event Data}
Assuming the ground is a perfect 2D horizontal plane, the vehicle can move ideally in a straight line under a constant steering command.
Thus, the heading direction (i.e., direction of linear velocity) can be well approximated by the normalized translation vector between any two successive time instants.
Although there exist several ways for relative pose estimation using event data as input, such as by fitting a homography \cite{Gallego18cvpr, huang2023progressive}, they are typically computational expensive and require the camera to observe a planar scene.
Instead, we leverage a feature-based method for relative pose estimation in standard vision, which calculates the essential matrix from a set of feature correspondences.
\begin{figure}[t!]
  \centering
  \subfigure[t][\small{Intensity image.}]{
  \includegraphics[width=0.3\columnwidth]{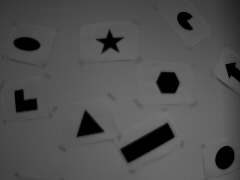}
  \label{fig:intensity}}\!\!
  \subfigure[t][\small{TS~\cite{Lagorce17pami}.}]{
  \includegraphics[width=0.3\columnwidth]{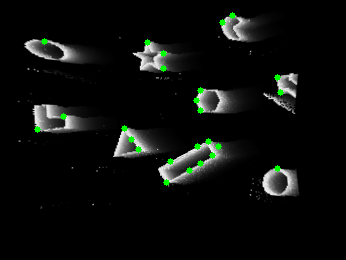}
  \label{fig:TS}}\!\!
  \subfigure[t][\small{DiST~\cite{Kim2021iccv}.}]{
  \includegraphics[width=0.3\columnwidth]{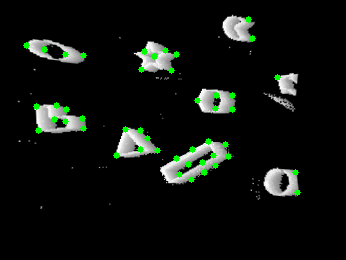}
  \label{fig:DiST}}\!\!
  \subfigure[t][\small{SILC~\cite{Manderscheid19cvpr}.}]{
  \includegraphics[width=0.3\columnwidth]{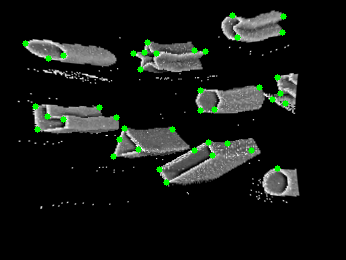}
  \label{fig:SILC}}\!\!
  \subfigure[t][\small{TOS~\cite{glover2021luvharris}.}]{
  \includegraphics[width=0.3\columnwidth]{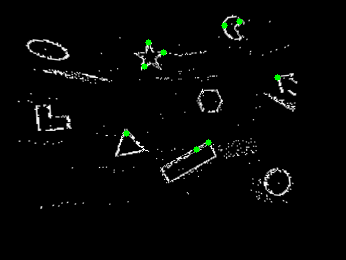}
  \label{fig:TOS}}\!\!
  \subfigure[t][\small{Ours.}]{
  \includegraphics[width=0.3\columnwidth]{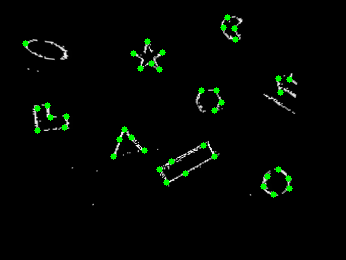}
  \label{fig:TOS+TS}}
  
  \caption{Various image-like representations of event data~\cite{Mueggler17ijrr} and results of corner detection~\cite{alzugaray18ral}.
  (a) is the intensity image, which is only for visualization.
  (b)-(e) are state-of-the-art representations of event data.
  (f) shows our design.}
  \label{fig:event representation}
  \vspace{-1cm}
\end{figure}

Detection and matching of event-based features (e.g., corners) have to be carried out on image-like representation of event data.
Assuming the photometric pattern of the environment is stable, the output of an event-based camera is dependent on the relative speed with respect to the scene.
Thus, an ideal representation, on one hand, is supposed to be invariant to speed, and, on the other hand, can be rendered (refreshed) efficiently.
To discover the optimal representation, we investigate a number of existing representations, including Time Surface (TS)~\cite{Lagorce17pami}, Discounted Sorted Timestamp Image (DiST)~\cite{Kim2021iccv}, Speed Invariant Time Surface (SILC)~\cite{Manderscheid19cvpr}, and Threshold-Ordinal Surface (TOS)~\cite{glover2021luvharris}.
TS is a 2D representation where each pixel stores a single time value, e.g., the timestamp of the most recent event at that pixel~\cite{Delbruck08issle}.
Using an exponential decay kernel~\cite{Lagorce17pami}, a TS (Fig.~\ref{fig:TS}) emphasizes recent events over past events and has been proved discriminative in pattern recognition tasks.
Similarly, DiST~\cite{Kim2021iccv} (Fig.~\ref{fig:DiST}) aims at preserving semantic profiles for object recognition under the camera's motion, and it does a great job in noise suppression.
These two representations are, however, not invariant to the event camera's speed.
The appearance similarity, even at two close views, will not be preserved in the presence of speed variation, and thus, no guarantee for the success of short-baseline feature matching.
To circumvent this issue, some hand-crafted representations (e.g., SILC~\cite{Manderscheid19cvpr} (Fig.~\ref{fig:SILC}) and TOS~\cite{glover2021luvharris} (Fig.~\ref{fig:TOS})) are designed to keep the 2D spatial gradient of moving edges constant under a variation of camera's speed.
This is basically achieved by continuously assigning the most recently firing coordinates to the maximum value, and reducing the magnitude of the adjacent area by a distance-related quantity.
The main issue of these two speed-invariant representations is about the relatively low signal-to-noise ratio, which is largely due to the fact that historical information is not recycled in time.

To hold the speed-invariant property while maintaining a good signal-to-noise ratio, we propose a novel representation by combining TS and TOS.
Given a TS and a TOS rendered by the same time, our representation is specifically obtained by performing a logical AND operation on corresponding pixels of the two maps.
As seen in Fig.~\ref{fig:TOS+TS}, the resulting representation inherits the speed-invariant property from TOS; meanwhile it is much cleaner than the original TOS due to the exponential decay kernel used in the TS.
We apply Arc$^\star$~\cite{alzugaray18ral} for corner detection and BRIEF~\cite{calonder2010brief} for feature description and matching.
As seen from Fig.~\ref{fig:TS} to Fig.~\ref{fig:TOS+TS}, more true-positive corners can be detected on the resulting representation, and thus, it is beneficial to solving the essential matrix in the following.

To calculate the essential matrix, we implement a five-point algorithm~\cite{nister2004efficient} inside a RANSAC scheme.
We notice that the essential matrix is a skew-symmetric matrix due to the absence of rotation, and therefore, the resulting translation vector can be straightforwardly retrieved from it.
Finally, the resulting heading direction from event data, denoted by $\bfv_{e}$, can be approximated by the normalized translation vector.
\subsection{Acquiring Body-Frame Velocity from Kinematic Model}
\label{subsec:Acquiring Body-Frame Velocity from Kinematic Model}

The ground vehicle used in this work is an all-wheel-steering mobility system as illustrated in Fig.~\ref{fig:eye catcher}.
Different from commonly seen non-holonomic counterparts (e.g., Ackermann mobility systems \cite{howard2007optimal}), an all-wheel-steering vehicle is capable of moving in any direction by simply steering all wheels into a certain angle.
Therefore, the direction of the linear velocity is not always along the X-axis of the body frame, and such a kinematic property enables us to obtain linear velocity measurements in various directions in the vehicle's body frame.

\begin{figure}[t]
    \centering
    \includegraphics[width=0.75\linewidth]{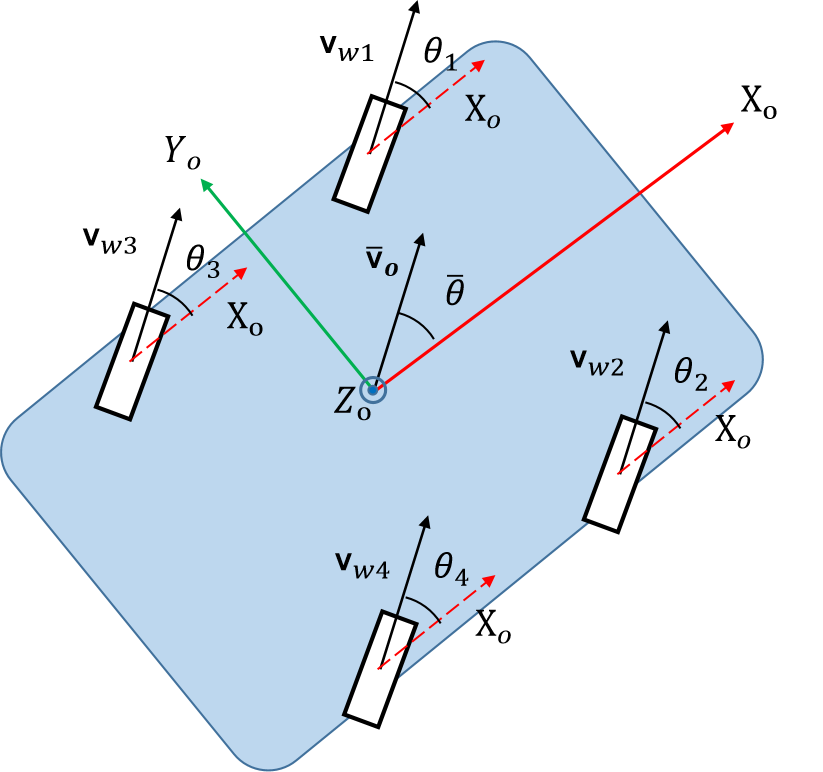}
    \caption{Kinematic model of the all-wheel-steering mobile platform.
    Pure linear motion at arbitrary direction can be realized by controlling all wheels with same steering angle and identical rotating speed.}
    \vspace{-0.8cm}
    \label{fig:kinematic model}
\end{figure}

The vehicle's body frame is defined as the coordinate system in Fig.~\ref{fig:kinematic model}.
Note that the location of the origin can be set anywhere, which does not affect the extrinsic rotation result.
What matters is the definition of the body frame's orientation.
The X-axis is defined to be with the longitudinal direction, and the Y-axis is the lateral direction.
The remaining Z-axis can be obtained using the right-hand rule.
The four wheels are actuated independently, and the corresponding odometer can report two states, i.e., the steering angle $\theta_i$ and the wheel speed $v_{wi}$.
The linear velocity of the vehicle can be simply derived from
\begin{equation}
\bfv_{o} =
\left [ \begin{matrix}
\bar{v}_w \\
0 \\
0 \\
\end{matrix} \right]
\left [ \begin{matrix}
cos(\bar{\theta}) & -sin(\bar{\theta}) & 0\\
sin(\bar{\theta}) & cos(\bar{\theta}) & 0 \\
0 & 0 & 1
\end{matrix} \right], 
\label{eq:robot_velocity_bearing_vector}
\end{equation}
where $\bfv_{o}$ denotes the linear velocity (in metric scale) represented in the body frame $o$, 
and $\bar{v}_w$ and $\bar{\theta}$ the average speed and steering angle of the four wheels.
Note that we use only the direction of linear velocity in the following extrinsic calibration.
With a slight abuse of mathematical notation, for brevity, we hereafter denote the direction by $\bfv_{o}$.

\subsection{Extrinsic Calibration via Correlation Maximization}
\label{subsec:Extrinsic Calibration via Correlation Maximization}
The above velocity direction estimates, $\bfv_{e}$ and $\bfv_{o}$, are linearly correlated through the extrinsic rotation $\Rot_{oe}$, as
\begin{equation}
\bfv_{o} = \Rot_{oe}\bfv_{e}.
\label{eq:linear correlation}
\end{equation}
Obviously, the minimal problem to solve $\Rot_{oe}$ requires two pairs of velocity direction estimates, which must be not parallel.
In the absence of temporal offset, the extrinsic rotation calibration can be simply solved as a 3-D point registration problem using the closed-form solution in \cite{arun1987least}.
However, simply neglecting the temporal offset may lead to inaccurate spatial calibration.
Thus, we need a method to determine the temporal offset in the presence of unknown extrinsic rotation.

CCA is an effective tool for evaluating the linear correlation between two random vectors (i.e., discrete samples of two ``synchronized'' signals).
It is typically used to recover the underlying linear transformation from signals' statistical profiles, such as covariances.
Assuming that we have collected two sets of ``synchronized'' velocity direction estimates, denoted by $\cV_{o} \doteq \{\bfv_{o}(t_i)\}_{i=1}^{N_{o}}$ and $\cV_{e} \doteq \{\bfv_{e}(t_i)\}_{i=1}^{N_e}$,
the cross-covariance and auto-covariance can be approximately calculated, in the presence of temporal offset $t_d$, by
\begin{align}
\Sigma_{v_{o}v_{e}}(t_d) &\approx \frac{1}{N-1}\sum_{i=0}^{N-1}(\bfv_{o}(t_i) - \bar{\bfv}_{o})(\bfv_{e}(t_i + t_d) - \bar{\bfv}_{e}^{\prime})^{\ttT},\nonumber\\ 
\Sigma_{v_{o}v_{o}}(t_d) &\approx \frac{1}{N-1}\sum_{i=0}^{N-1}(\bfv_{o}(t_i) - \bar{\bfv}_{o})(\bfv_{o}(t_i + t_d) - \bar{\bfv}_{o}^{\prime})^{\ttT},\nonumber\\
\Sigma_{v_{e}v_{e}}(t_d) &\approx \frac{1}{N-1}\sum_{i=0}^{N-1}(\bfv_{e}(t_i) - \bar{\bfv}_{e})(\bfv_{e}(t_i + t_d) - \bar{\bfv}_{e}^{\prime})^{\ttT},
\label{eq:cross-variance and auto-variance}
\end{align}
where $t_i$ refers to the sampling time, $N$ the number of samples, $\bar{(\cdot)}$ the mean of original samples, and $\bar{(\cdot)}^{\prime}$ the mean of temporal-offset compensated samples.

For multi-variable random vectors, the underlying linear transformation is made from several linear combination pairs.
In our case, the linear combination pairs are $\bfs_i \leftrightarrow \bfr_i$ $(i = \{1, 2, 3\})$, which represent respectively the basis vectors of the 3-by-3 identity matrix and the transpose of the extrinsic rotation matrix, namely $\bfI_{3\times3} = [\bfs_{1} \vert \bfs_{2} \vert \bfs_{3}]$ and $\Rot_{oe}^{\ttT} = [\bfr_{1} \vert \bfr_{2} \vert \bfr_{3}]$.
Each column vector $\bfr_i$ can be determined by maximizing the correlation coefficient  
\begin{align}
\begin{split}
\rho_{i} &\doteq Corr(\bfs_{i}^{\ttT}\bfv_{o}, \bfr_{i}^{\ttT}\bfv_{e}) \\ 
&= \frac{\bfs_{i}^{\ttT}\Sigma_{v_{o}v_{e}}\bfr_i}{\sqrt{\bfs_{i}^{\ttT}\Sigma_{v_{o}v_{o}}\bfs_{i}}\sqrt{\bfr_{i}^{\ttT}\Sigma_{v_{e}v_{e}}\bfr_{i}}}, i \in \{1,2,3\},
\label{eq: correlation coefficient}
\end{split}
\end{align}
With the three canonical correlation coefficients, the so-called trace correlation between the two set of estimates is further defined as
\begin{equation}
r(\cV_{o}, \cV_{e}) = \sqrt{\frac{1}{3}\sum_{i=1}^3\rho_i^2}
=\sqrt{\frac{1}{3}\mathrm{Tr}(\Sigma_{v_{o}v_{o}}^{-1}\Sigma_{v_{o}v_{e}}\Sigma_{v_{e}v_{e}}^{-1}\Sigma_{v_{e}v_{o}})},
\label{eq:trace correlation}
\end{equation}
where $\mathrm{Tr}(\cdot)$ denotes the trace of an input matrix.
It is also a normalized measurement that evaluates the correlation between two signals.
One of the great properties of trace correlation is that it is invariant to underling linear transformation between two signals, such as scaling, rotation and translation.
In our case, hence, we have $r(\cV_{o}, \Rot_{oe}\cV_{e}) = r(\cV_{o}, \cV_{e})$.
This property decouples the impact of the temporal offset on the trace correlation from the unknown spatial extrinics.
Therefore, the optimal $t_d$ can be independently obtained by maximizing eq.~\ref{eq:trace correlation}.

Once the temporal offset is compensated in the data, the extrinsic rotation $\Rot_{oe}$ can be simply worked out as a by-product of the above CCA process \cite{qiu2020real}.
Using the covariance matrices obtained in the CCA process, the extrinsic rotation can be calculated by
\begin{equation}
\Rot_{oe}^{\ttT} = \bfU
\left[
\begin{matrix}
1 & 0 & 0 \\
0 & 1 & 0 \\
0 & 0 & \mathtt{det}(\bfU\bfV^{\ttT})
\end{matrix}
\right]
\bfV^{\ttT},
\label{eq:rigid registration on statistic shape}
\end{equation}
where $\bfU$ and $\bfV$ are obtained from the following singular value decomposition (SVD)
\begin{equation}
\Sigma_{v_{e}v_{e}}^{-1}\Sigma_{v_{e}v_{o}} = \bfU \boldsymbol{\Sigma} \bfV^{\ttT}.
\label{eq:svd decomposition}
\end{equation}
The determinant operation in Eq.~\ref{eq:rigid registration on statistic shape} is to guarantee the resulting rotation matrix is not a reflection.
However, this quick solver cannot be straightforwardly applied to our case.
The kinematic characteristics of the ground vehicle gives rise to zero-velocity measurements in $z$ axis of the body frame.
A big condition number is witnessed for $\Sigma_{v_{e}v_{e}}^{-1}\Sigma_{v_{e}v_{o}}$, and thus, the decomposition result is always numerically unstable.
Although this issue can be more or less mitigated by adding an additional perturbation in the non-stimulated dimension, accurate results are still not guaranteed as seen in our experiments.

\begin{algorithm}[b!]
\caption{Spatio-Temporal Calibration Method}
\label{alg:Temporal-Spatial calibration algorithm}

\begin{algorithmic}[1]
    \Require Linear velocity estimates $\cV_{o} = \{\bfv_{o,t_i}\}_{i=1}^{N_{o}}$, $\cV_{e} = \{\bfv_{e}(t_j)\}_{j=1}^{N_e}$ and parameter $t_{max}$   %
    \Ensure  Temporal offset $t_d^*$ and extrinsic rotation $\Rot_{oe}$   %

    \State \textbf{function} SpatioTemporalCalibration($\cV_{o}$,$\cV_{e}$,$t_{max}$)
    \State Initialize $\cV_{o}^* = \emptyset$, $r_{max} = -1$, and $t_d^* = 0$
    \For {each $t_d \in [-t_{max}, t_{max}]$ }
        \State $\cV_{o}^{t_d} = \emptyset$
        \For {$j = 1$ to $N_e$}
            \State $\bfv_{o}^{t_d}(t_j) = DataInterpolation(\cV_{o}, t_j + t_d)$
            \State $\cV_{o}^{t_d} \leftarrow \bfv_{o}^{t_d}(t_j)$ \Comment{Pushback operation}
        \EndFor
        
        \If{$r(\cV_{o}^{t_d}, \cV_{e}) \textgreater r_{max}$}
           \State $r_{max} = r(\cV_{o}^{t_d}, \cV_{e})$
           \State $\cV_{o}^* = \cV_{o}^{t_d}$, $t_d^* = t_d$
        \EndIf
    \EndFor
    \State Set $w_{i} = 1, i = 1,2,...,N_e$ %
	\While{$\Rot$ does not converge}
		\State $\mathbf{H} = \sum_{i=1}^{N_e}w_{i}\bfv_{o}^*(t_i) {\bfv_{e}}^{T}(t_i)$
    	\State $\mathbf{U}\mathbf{\Sigma}\mathbf{V^{T}} \leftarrow$ svd($\mathbf{H}$)
    	\State $\Rot = \mathbf{U}
                            \left[
                            \begin{matrix}
                            1 & 0 & 0 \\
                            0 & 1 & 0 \\
                            0 & 0 & \mathtt{det}(\mathbf{U}\mathbf{V}^{\ttT})
                            \end{matrix}
                            \right]
                            \mathbf{V}^{\ttT}$ \Comment{Reflection check, see~\cite{arun1987least}.}
        \State $w_{i} = \frac{1}{\max(\delta,\vert \bfv_{o}^*(t_i) - \Rot\bfv_{e}(t_i) \vert)}$ \Comment{$\delta$ is a small number}
    \EndWhile
\State \Return $\Rot, t_d^*$
\State \textbf{end function}
\end{algorithmic}
\end{algorithm}

To this end, we regard the two sets of linear velocity estimates as general ``point cloud'' on a unit sphere, and assort to a least-squares method~\cite{arun1987least} for registering two point sets.
Consequently, the extrinsic rotation can be calculated by maximizing
\begin{equation} 
\begin{split}
\mathcal{L} & = \sum_{i=1}^{N}\bfv_{o,t_i}^{T}\Rot_{oe}{\bfv_{e,t_i}} \\
& = \mathrm{Tr}(\sum_{i=1}^{N}\Rot_{oe}{\bfv_{e,t_i}}\bfv_{o,t_i}^{T}) = \mathrm{Tr}(\Rot_{oe}\bfH),
\end{split}
\label{eq:least squares problem2}
\end{equation}
where $\bfv_{o,t_i} \doteq \bfv_{o}(t_i)$, $\bfv_{e,t_i} \doteq \bfv_{e}(t_i)$, and $\mathbf{H} := \sum_{i=1}^{N}\bfv_{e,t_i}\bfv_{o,t_i}^{T}$.
Let the SVD of $\mathbf{H}$ be $\mathbf{U}\mathbf{\Lambda}\bfV^{T}$.
The resulting extrinsic rotation matrix is given as $\Rot_{oe} = \bfV\bfU^{T}$. 
To deal with noise and outliers, we apply an improved version of Arun's method \cite{zhou2016real}, which replaces the original sum of squares with sum of absolute values.
The resulting problem can be solved efficiently using an iteratively re-weighted least square method.
The whole spatio-temporal calibration method is summarized in Alg.~\ref{alg:Temporal-Spatial calibration algorithm}.

\begin{figure}[t!]
  \centering
  \subfigure[t][\small{Perfect arc trajectories.}]{
  \includegraphics[width=0.45\columnwidth]{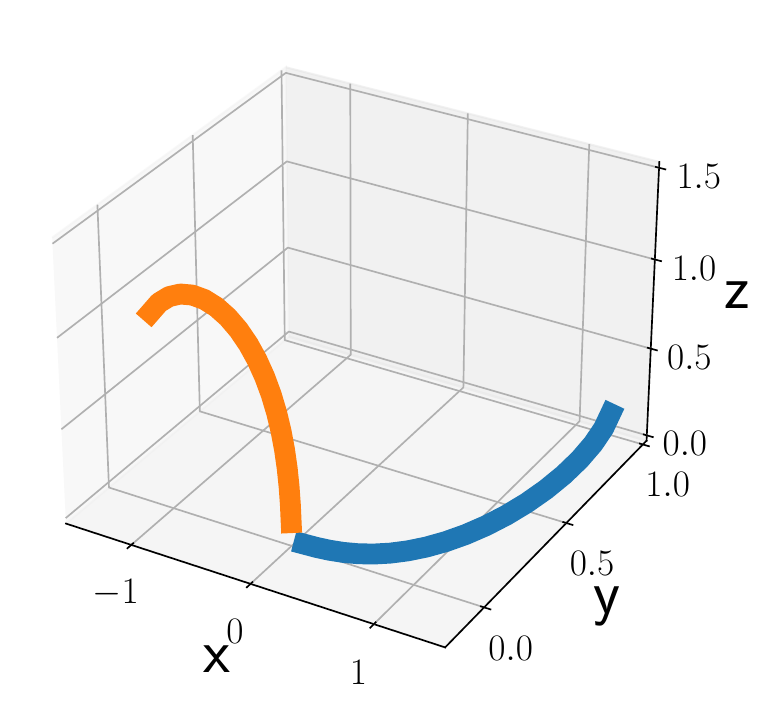}
  \label{subfig:arc_traj}}\!\!
  \subfigure[t][\small{Drifted arc trajectories.}]{
  \includegraphics[width=0.45\columnwidth]{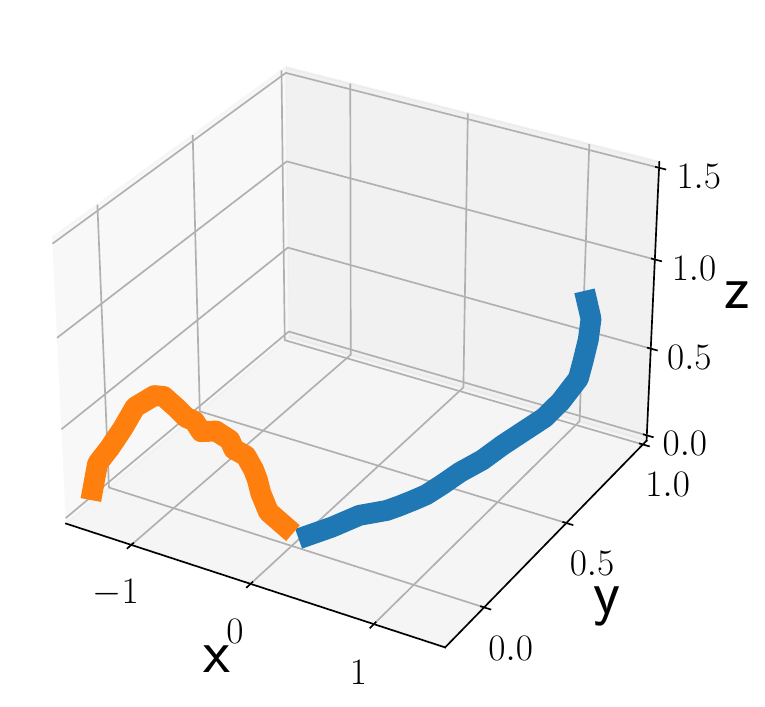}
  \label{subfig:arc_traj_drifted}}\!\!
  \subfigure[t][\small{Perfect polyline trajectories.}]{
  \includegraphics[width=0.45\columnwidth]{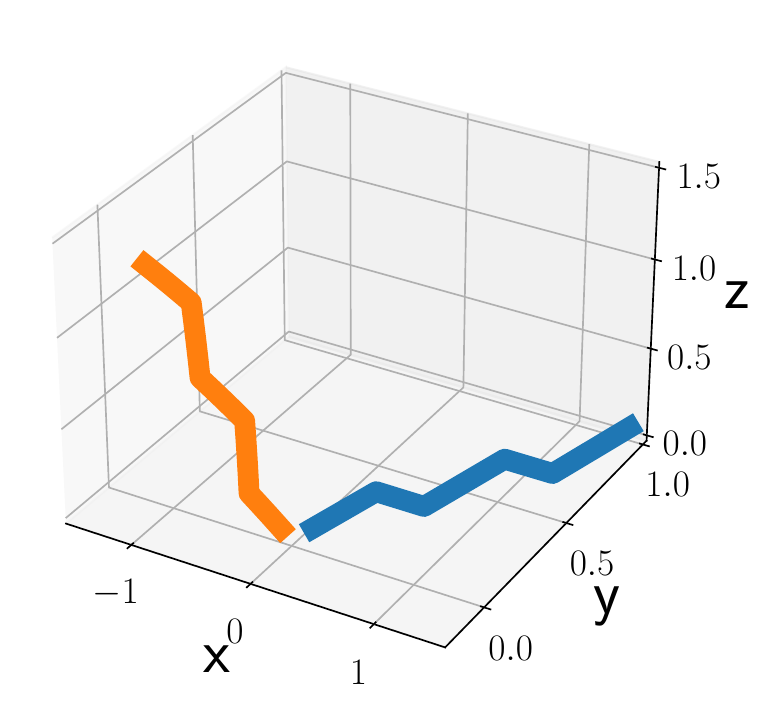}
  \label{subfig:poly_traj}}\!\!
  \subfigure[t][\small{Drifted polyline trajectories.}]{
  \includegraphics[width=0.45\columnwidth]{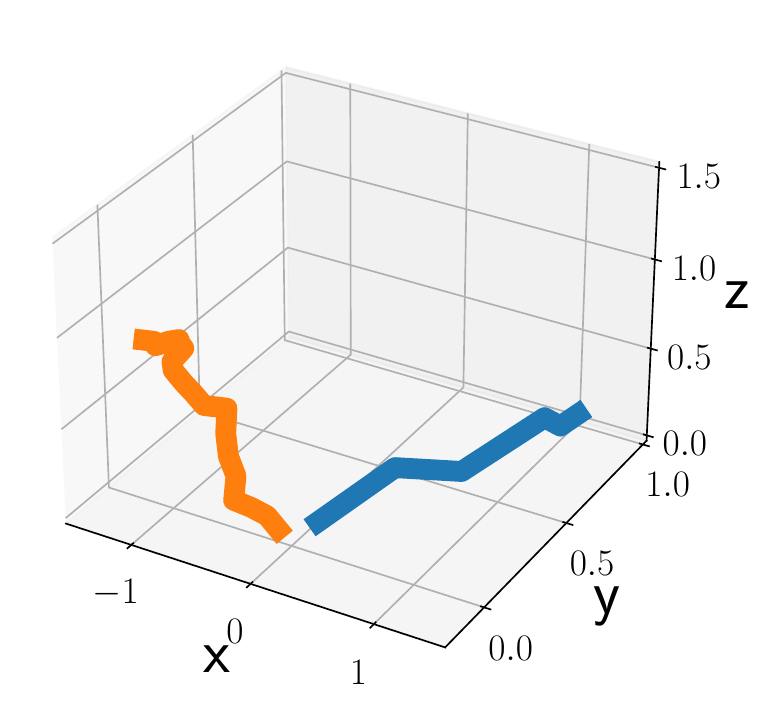}
  \label{subfig:poly_traj_drifted}}\!\!
  \caption{Simulation of perfect trajectories and corresponding drifted ones.
  Trajectories from the two heterogeneous sensors are expressed in a unified coordinate system.
  Different colors are used to distinguish trajectories from heterogeneous sources.}
  \label{fig:synthetic_trajecotry}
  \vspace{-1cm}
\end{figure}
\section{Experiments}
\setlength{\parskip}{-1pt}
\label{sec:evaluation}

In this section, we evaluate the proposed calibration method.
We first show that how synthetic data and real data for evaluation are generated and collected in Sec.~\ref{subsec:synthetic data} and Sec.~\ref{subsec:real data}, respectively.
Then we quantitatively evaluate the proposed method and compare against trajectory-alignment based pipelines in Sec.~\ref{subsec:evaluation metrics and results}, demonstrating the advantage and efficacy of our method.

\subsection{Generation of Synthetic Data}
\label{subsec:synthetic data}

In order to demonstrate the negative impact of drifted trajectories on trajectory-alignment based calibration methods, we generate two types of input trajectories to be aligned, which include an arc-shape trajectory (Fig.~\ref{subfig:arc_traj}) and a polyline-shape trajectory (Fig.~\ref{subfig:poly_traj}).
The perfect arc-shape trajectory can be determined solely by one parameter, i.e., the radius of the big circle.
The direction of instantaneous linear velocity at any way point on the arc can be determined as the tangential direction, while its magnitude is determined by the instantaneous angular velocity.
Without loss of generality, we simply use a constant angular velocity in each trial.
As for the perfect polyline-shape trajectory, the design parameters consist of the turning angle and the length of each segment.
Still, we assume these two parameters are constant for simplicity.
Therefore, the direction of instantaneous linear velocity at any way point (except for those corners) is identical to that of the corresponding segment, and its magnitude is set with a constant number.
\begin{figure}[t!]
  \centering
  \subfigure[t][\small{Finger-shape trajectory.}]{
  \includegraphics[width=0.45\columnwidth]{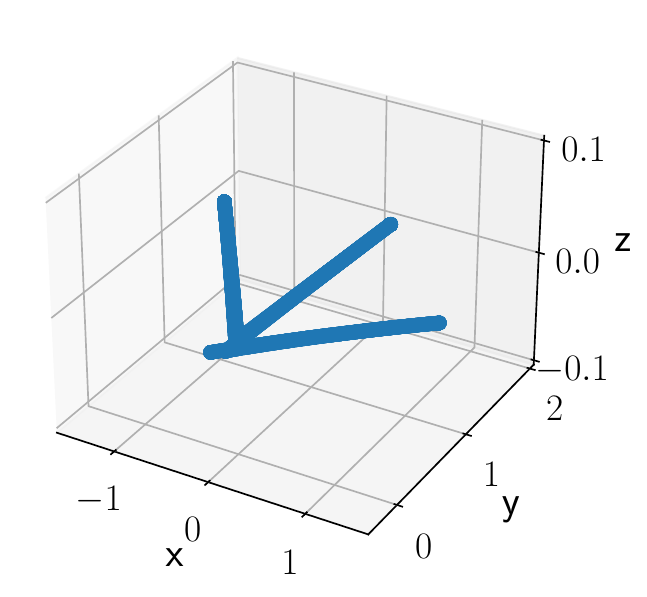}
  \label{subfig:real_data_traj}}\!\!
  \subfigure[t][\small{Linear velocities}]{
  \includegraphics[width=0.45\columnwidth]{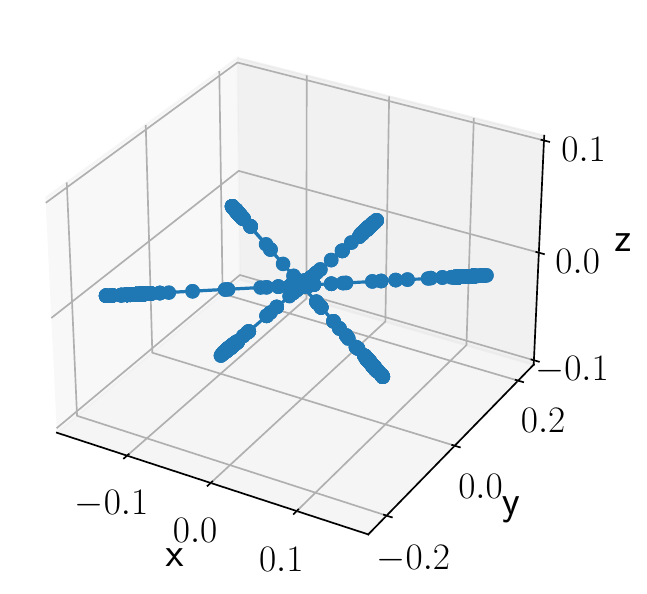}
  \label{subfig:real_data_vel}}\!\!
  \caption{An example of finger-shape trajectories carried out in the collection of real data. (a) The trajectory of three ``fingers''. (b) Corresponding linear velocity measurements.}
  \label{fig:real_world_traj_ours}
\end{figure}
\begin{figure}[t!]
  \centering
  \subfigure[t][\small{Image of the scene.}]{
  \includegraphics[width=0.45\columnwidth]{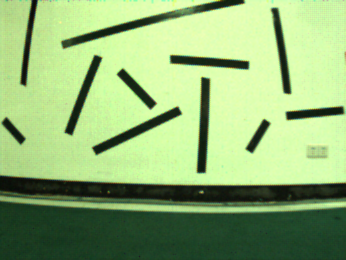}
  \label{subfig:real_data_raw}}\!\!
  \subfigure[t][\small{Image-like representation.}]{
  \includegraphics[width=0.45\columnwidth]{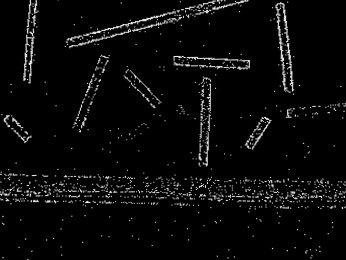}
  \label{subfig:real_data_tos2}}\!\!
  \caption{Illustration of visual information in the real data.
  (a) A sample RGB image of the scene for visualization only. 
  (b) Our image-like representation rendered by the same time using events collected.}
  \label{fig:real-data-representation}
  \vspace{-0.5cm}
\end{figure}
To simulate spatial drifts in the trajectories to be aligned (see Fig.~\ref{subfig:arc_traj_drifted} and Fig.~\ref{subfig:poly_traj_drifted}), we add Gaussian noises to the perfect linear velocities and obtain the drifted trajectory by dead reckoning according to the following simple kinematics:
\begin{equation}
\begin{aligned}
\hat \bfv_{o,i} &= \bfv_{o,i} + \boldsymbol{\epsilon}_o,      \\
\hat \bfv_{e,i} &= \Rot_{\texttt{E}} \bfv_{o,i} + \boldsymbol{\epsilon}_e,     \\
\bft_{o,i} &= \bft_{o,i-1} + \hat \bfv_{o,i-1} \cdot dt, \\
\bft_{e,i} &= \bft_{e,i-1} + \hat \bfv_{e,i-1} \cdot dt, \\
\end{aligned}
\label{eq:synthetic_data}
\end{equation}
where $\boldsymbol{\epsilon}_o$ and $\boldsymbol{\epsilon}_e$ denote the added Gaussian noises, and $\Rot_{\texttt{E}}$ the groundtruth extrinsic rotation.
To simulate temporal offset between data, we furthermore slide temporally one drifted trajectory by a certain time length with respect to the other.
As a result, we have temporally non-aligned and spatially perturbed kinematic information (i.e., linear velocities and way points) from two heterogeneous sources.
For extensive evaluation, we generate six groups of data for each type of trajectory with different design parameters, and in each group we propose 100 trials with random temporal offset and extrinsic rotation.

\subsection{Collection of Real Data}
\label{subsec:real data}

We also collect real data using an event camera mounted on a non-holonomic all-wheel-steering mobile platform.
By periodically adjusting the steering angle of all the wheels, we can obtain finger-shape trajectories which are friendly to our calibration method.
An example of three-finger-shape trajectory and the corresponding linear velocity information are illustrated in Fig.~\ref{fig:real_world_traj_ours}.
The measurement of linear velocity (i.e., $\bfv_{o}$) can be obtained according to Eq.~\ref{eq:robot_velocity_bearing_vector} using as input the control commands of steering angle and wheel speed. 
We collect four groups of data with various number of ``fingers'', aiming to show that the more directions the data contain the more accurate the calibration result. 
An illustration of visual data is shown in Fig~\ref{fig:real-data-representation}.
As for the groundtruth extrinsic rotation, we manage to set the event camera at a desired orientation with the help of two independent laser level meters.

\subsection{Evaluation Metrics and Results}
\label{subsec:evaluation metrics and results}

We evaluate our method using the above datasets, and compare against existing pipelines listed in the following:
\begin{itemize}
    \item CGOC: A globally optimal solution method using quadratically constrained quadratic programs (QCQPs) proposed in~\cite{giamou2019certifiably}.
    \item Hand-eye: An implementation based on  Eq.~\ref{eq: hand-eye calibration energy function}.
    \item VC: The proposed velocity-correlation based method.
    \item VC-woTA: The proposed method without temporal alignment.  
\end{itemize}

\begin{figure}[t!]
  \centering
  \subfigure[t][\small{Results on simulated arc-shape trajectory.}]{
  \includegraphics[width=0.9\columnwidth]{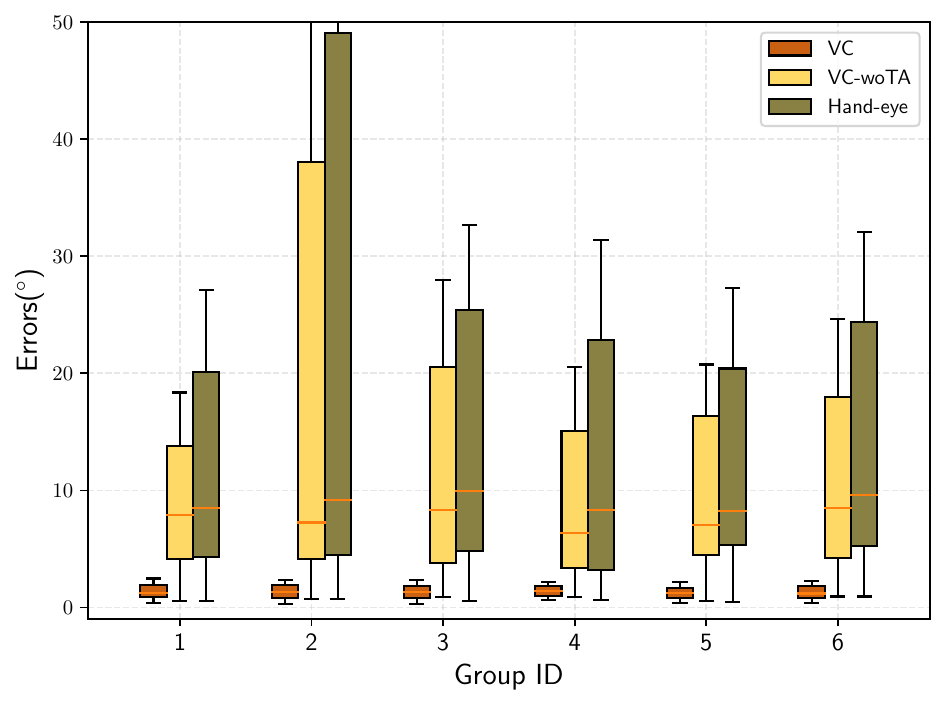}
  \label{fig:result_arc_traj}}\!\!
  \subfigure[t][\small{Results on simulated polyline-shape trajectory.}]{
  \includegraphics[width=0.9\columnwidth]{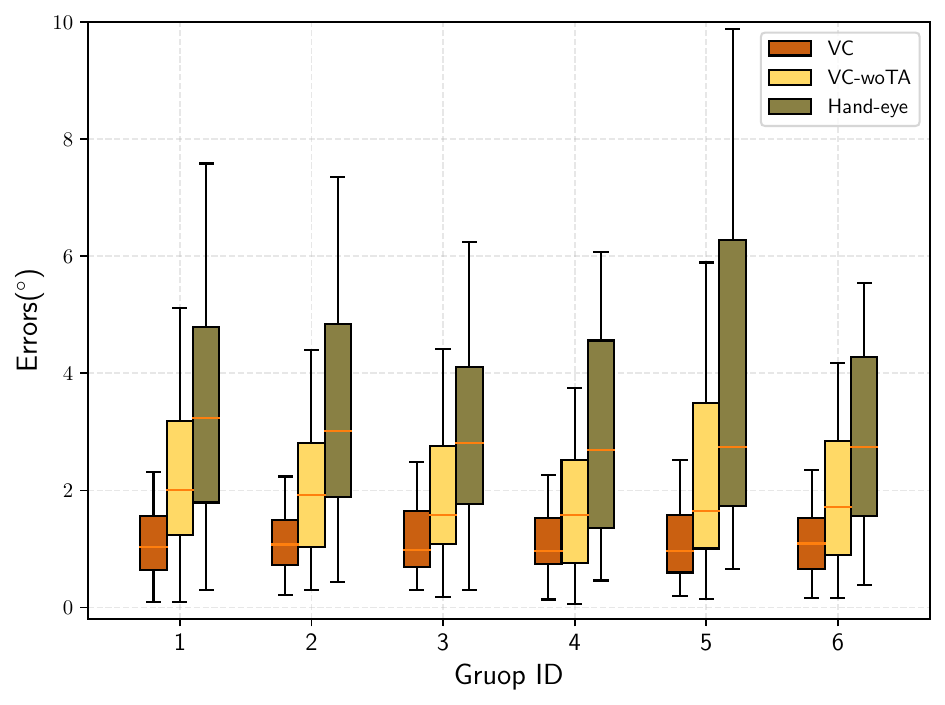}
  \label{fig:result_polyline_traj}}\!\!
  \caption{Illustration of statistics of the calibration error.}
  \label{fig:temporal_offset_result}
 \vspace{-0.8cm}
\end{figure}
\noindent The metric used for quantitative evaluation of extrinsic rotation is defined as
\begin{equation}
e = arccos((\mathrm{Tr}(\Rot_{g}^T \Rot_{est}) - 1)/2)*180/\pi,
\label{eq:rotation_metric}
\end{equation}
where $\Rot_{\tt{est}}$ and $\Rot_{g}$ are the estimation result and the groundtruth orientation, respectively.

Using the synthetic dataset, we calculate the calibration error according to Eq.~\ref{eq:rotation_metric} and illustrate the error's statistics using box plot.
As seen in Fig.~\ref{fig:temporal_offset_result}, our methods (VC and VC-woTA) outperform Hand-eye calibration method in terms of accuracy.
Specifically, we clearly observe that VC outperforms VC-woTA.
This indicates that temporal calibration is vital in the calibration task involving heterogeneous sensors.
Furthermore, VC-woTA outperforms Hand-eye, and this result validates our point: Trajectory-alignment based methods suffer from drifted trajectories while methods based on measurements of instantaneous first-order kinematics (e.g., linear velocity in our case) can work.
Besides, we find that the calibration results of the polyline-shape dataset are typically more accurate than those of the arc-shape dataset.
This is because the temporal offset has less effect on the polyline-shape dataset than on the arc-shape dataset.

To evaluate on real data, we first recover trajectories from data of the event camera and the odometers, respectively.
The trajectory from event data is estimated using EVO~\cite{Rebecq17ral}, a state-of-the-art visual odometry pipeline for a monocular event camera.
The direction of linear velocity from event data is obtained using our method discussed in Sec.~\ref{subsec:Determining Heading Direction from Event Data}.
Meanwhile, the linear velocity from odometers is obtained as shown in Sec.~\ref{subsec:Acquiring Body-Frame Velocity from Kinematic Model}, and the trajectory is obtained by dead reckoning.
Since the estimates from the heterogeneous sensors are typically not synchronized, we obtain data association (synchronized and temporally-non-aligned) by data interpolation.
Compared to the evaluation on synthetic dataset, we introduce another trajectory-alignment based solution \cite{giamou2019certifiably}, which can return globally optimal extrinsic estimates.
As seen in Table.~\ref{tab:real_data}, the best results are highlighted in bold, and the conclusion is consistent with that in the evaluation on synthetic data.
Our methods (VC and VC-woTA) outperform trajectory-alignment based methods, and more specifically, VC outperforms VC-woTA.
Besides, we see a clear trend in the last column of Table.~\ref{tab:real_data} that the diversity of linear velocity direction (i.e., the number of ``fingers'') would promote the accuracy of calibration results.
This is due to the fact that the more directions are considered, the more distinctive the kinematic profile becomes.

 \begin{table}[t]
     \caption{Calibration errors on real data ($^\circ$)}%
     \centering%
     \begin{tabular}{cccccc}%
         \toprule%
         \#Fingers  & CGOC~\cite{giamou2019certifiably} & Hand-eye & VC-woTA & VC   \\
         \midrule%
         2 & 86.22 & 35.59   & 6.29 & \textbf{5.97} \\
         3 & 38.98 & 5.82   &  2.02 & \textbf{1.98}\\
         4 & 7.96 & 9.19 & 2.11 & \textbf{2.01}\\
         5 & 6.21  & 7.20 & 1.82 & \textbf{1.68} \\
         \bottomrule%
     \end{tabular}
     \label{tab:real_data}
     \vspace{-1cm}
\end{table}

\section{Conclusion}
\label{sec:conclusion}

This paper provides a novel solution to the problem of spatio-temporal calibration for 
omni-directional vehicle mounted event cameras.
We argue that trajectory-alignment based methods suffer from drifts in any of the input trajectories.
To this end, we propose a two-step method that establishes correlation on first-order dynamics, namely instantaneous linear velocity.
In the first step, the optimal temporal offset is estimated by maximizing a correlation measurement invariant to the unknown extrinsic rotation.
In the second step, we regard directions of linear velocity estimates as general point clouds on a unit sphere and calculate the extrinsic rotation matrix via a point cloud registration process, which is accurately and robustly solved using an iteratively re-weighted least squares method.
Experiments on both synthetic data and real data demonstrate the efficacy of the proposed calibration method.
Finally, we hope this work inspires new research in the topic of multi-sensor calibration that involves event-based cameras.

\bibliographystyle{IEEEtran} %
\bibliography{myBib}

\end{document}